%% file: Main.tex
\documentclass[10pt,twocolumn,letterpaper]{article}

\usepackage[pagenumbers]{cvpr} 

\usepackage{graphicx}
\usepackage{amsmath}
\usepackage{amssymb}
\usepackage{booktabs}
\usepackage{comment}
\newcommand{\myparagraph}[1]{\vspace{4pt}\noindent\textbf{#1}}
\usepackage{multirow}
\usepackage{adjustbox}
\newcommand*\samethanks[1][\value{footnote}]{\footnotemark[#1]}
\usepackage[accsupp]{axessibility}
\usepackage[pagebackref,breaklinks,colorlinks]{hyperref}

\usepackage[capitalize]{cleveref}
\crefname{section}{Sec.}{Secs.}
\Crefname{section}{Section}{Sections}
\Crefname{table}{Table}{Tables}
\crefname{table}{Tab.}{Tabs.}

\def\cvprPaperID{5} 
\def\confName{CVPR}
\def\confYear{2022}

\usepackage{my_ref}

\begin{document}

\definecolor{ForestGreen}{RGB}{34,139,34}
\newcommand{\tav}[1]{{\color{ForestGreen}{T: #1}}}

\definecolor{CarloOrange}{RGB}{255,125,30}
\newcommand{\car}[1]{{\color{CarloOrange}{C: #1}}}

\title{Augmentation Invariance and Adaptive Sampling in \\Semantic Segmentation of Agricultural Aerial Images}

\author{
Antonio Tavera\thanks{Equal contribution}\ $^{1}$, Edoardo Arnaudo\samethanks\ $^{1,2}$, Carlo Masone$^{3}$, Barbara Caputo$^{1}$\\
$^1$Politecnico di Torino, Turin, Italy \\$^2$LINKS Foundation, Turin, Italy \\$^3$ Consorzio Interuniversitario Nazionale per l’Informatica, Rome, Italy \\
$^1${\tt\small \{first.last\}@polito.it} \\
$^2${\tt\small \{first.last\}@linksfoundation.com} \\
}
\maketitle

\input{Sections/0-Abstract}

\input{Sections/1-Introduction}

\input{Sections/2-RelatedWork}

\input{Sections/3-Method}

\input{Sections/4-Experiments}

\input{Sections/5-Ablation}

\input{Sections/6-Conclusion}

{\small
\bibliographystyle{ieee_fullname}
\bibliography{egbib}
}

\end{document}

%% file: Sections/0-Abstract.tex
\begin{abstract}
In this paper, we investigate the problem of Semantic Segmentation for agricultural aerial imagery. We observe that the existing methods used for this task are designed without considering two characteristics of  the aerial data: (i) the top-down perspective implies 
that the model cannot rely on a fixed semantic structure of the scene, because the same scene may be experienced with different rotations of the sensor; (ii) there can be a strong  imbalance in the distribution of semantic classes because the relevant objects of the scene may appear at extremely different scales (e.g., a field of crops and a small vehicle).
We propose a solution to these problems based on two ideas: (i) we use together a set of suitable augmentation and a consistency loss to guide the model to learn semantic representations that are invariant to the photometric and geometric shifts typical of the top-down perspective (Augmentation Invariance); (ii) we use a sampling method (Adaptive Sampling) that selects the training images based on a measure of pixel-wise distribution of classes and actual network confidence. 
With an extensive set of experiments conducted on the Agriculture-Vision dataset, we demonstrate that our proposed strategies improve the performance of the current state-of-the-art method.~\footnote{Code can be found at: \href{https://github.com/taveraantonio/AIAS}{https://github.com/taveraantonio/AIAS}.}.

\end{abstract}

%% file: Sections/1-Introduction.tex
\begin{figure}[!t]
\begin{center}
\includegraphics[width=1.0\columnwidth]{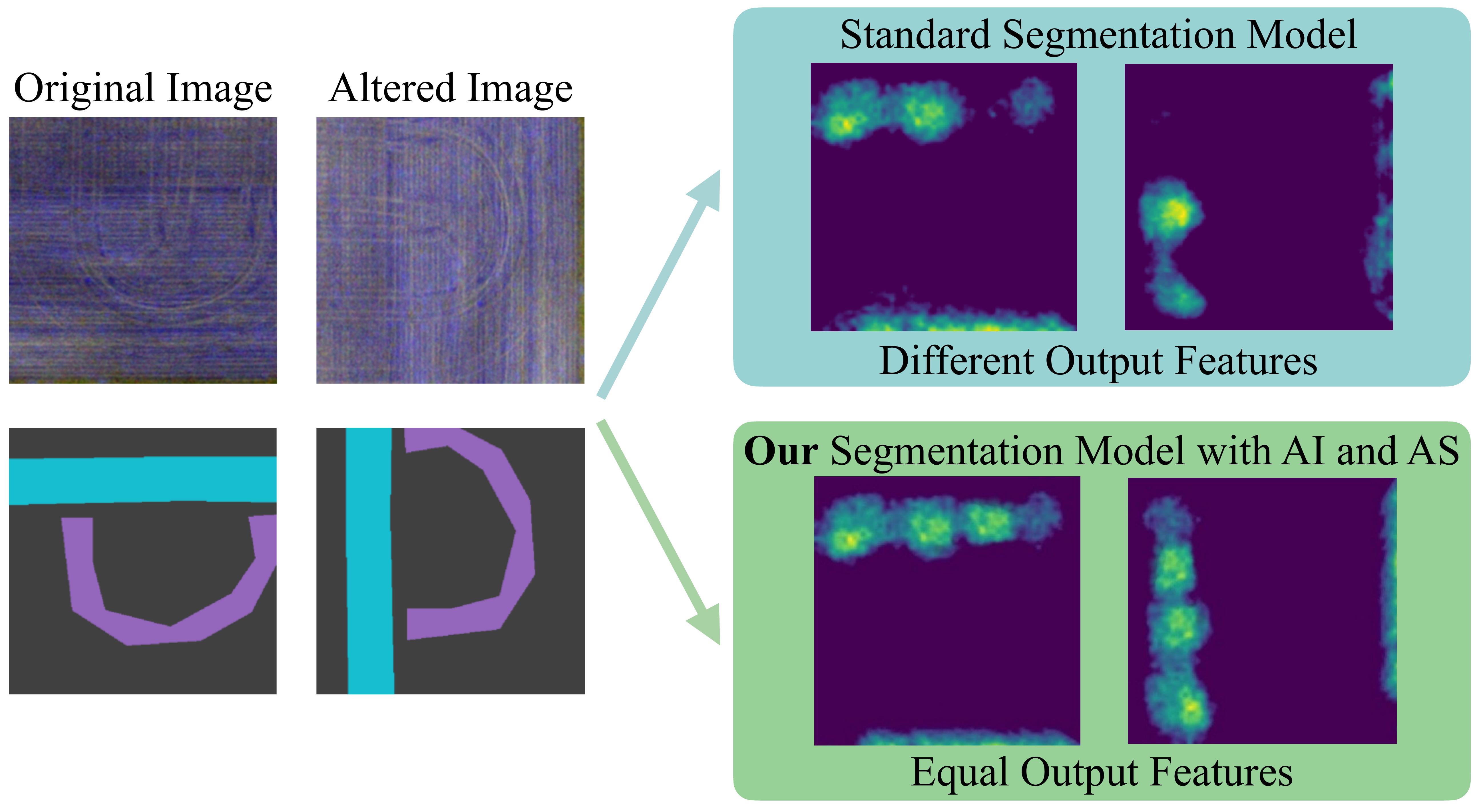}
\end{center}
\vspace{-10pt}
\caption{A semantic segmentation model that is not designed to expect changes in points of view, may produce different output features for the same image when seen from different angles. Our technique, on the other hand, makes the model invariant to these viewpoint shifts, encouraging the model to learn more robust representations.
}
\label{fig:teaser}
\end{figure}

\section{Introduction}

Semantic segmentation, i.e., the task of classifying each pixel of an image into a preset taxonomy of semantic categories, is a fundamental research problem in computer vision and a key technology in many real-world applications. Among these applications, the environmental monitoring from remote aerial images has grown considerably in recent years, with examples of categorization of land cover \cite{ref_deepglobe, ref_loveda}, delineation of wildfire \cite{farasin_doublestep}, and identification of deforested regions \cite{andrade_deforestation}.
In this task, as in other computer vision problems, deep learning models have demonstrated promising results, thanks also to an increasing availability of open datasets and large scale collections of aerial images \cite{ref_landcover_ai, ref_loveda}.
However, the majority of these deep learning models were originally designed for other use-cases, such as self-driving vehicles \cite{semseg_survey} and medical imaging \cite{ronneberger_unet}, and then transferred to the aerial domain without considering its specific characteristics. In particular, we find two peculiarities that set the task of aerial segmentation apart from its autonomous driving counterpart:

\myparagraph{Top-down perspective:}
In remote sensing the images are collected with a top-down perspective, i.e., from a camera mounted on an aircraft and pointed towards the ground. This remote perspective implies not only a lack of depth and reference points in the pictures, but it also allows to capture the same scene with arbitrary rotations around the vertical axis (see \cref{fig:teaser}). Thus, whereas in autonomous driving datasets \cite{cityscapes, idda} the model is bound to experience a well structured organization in the semantic elements of the scene (e.g., the road is expected at the bottom of the image, the sky on the top), this is not true in aerial imagery.

\myparagraph{Extreme class imbalance}
Although the problem of class imbalance in class-wise pixel distributions is typical of semantic segmentation \cite{Tavera_2022_WACV}, in aerial images this is brought to an extreme because the entities to be recognized range from small vehicles to large natural biomes.

We argue that a semantic segmentation model that is designed to account for these characteristics of the aerial setting can be more effective at the task. Thus, we propose a solution based on two ideas:
Augmentation Invariance (AI) and Adaptive Sampling (AS). 
The first one uses augmentations to guide the model to learn representations that are invariant to shifts in appearance and perspective (e.g., rotations around the vertical axis, as shown in \cref{fig:teaser}). 
The second is intended to  regularize the training of underrepresented classes by adaptively sampling the training images according to the distribution of pixels and the actual network confidence. 
These two modules cooperate in an end-to-end training stage. 

Summarizing, the contributions of this paper are:

\begin{itemize}
    \item An Augmentation Invariance technique that is tailored to handle the specific challenges given by the perspective in the aerial data and to help the model to separate semantic information from appearance.
    \item An Adaptive Sampling approach to address the problem of class imbalance, by dynamically sampling training data based on the current network confidence and the global, pixel-wise class distribution.
    \item An extensive set of experiments on the Agriculture Vision dataset \cite{ref_agrivision}, which is the only agricultural aerial dataset with several semantic classes and complexity. We show what happens when only RGB images are used for training, as well as when NIR data is exploited. Furthermore, an exhaustive ablation study examines the impact of all the solutions introduced. The code will be made available in order to encourage research.
\end{itemize}

%% file: Sections/2-RelatedWork.tex
\section{Related Work}

\subsection{Semantic Segmentation}
There is a flourishing literature on semantic segmentation, mostly pertaining different network architectures and techniques to capture the global context of a scene. Methods such as FCN \cite{long_fcn} include only convolutional layers and use skip connections to incorporate semantic and appearance information from deep and shallow layers, respectively. Most common segmentation models, such as U-Net \cite{ronneberger_unet}, HRNet \cite{sun_hrnetv1} and HRNetV2 \cite{wang_hrnetv2}, use an encoder-decoder structure to extract objects and image context at different scales. Multi-scale approaches are also used in solutions such as FPN \cite{lin_fpn}, UperNet \cite{xiao_upernet} and PSPNet \cite{zhao_pspnet} to better condition the global context of a scene. DeepLab V2 \cite{chen_deeplabv2} and V3 \cite{chen_deeplabv3} use the dilation parameter of convolutional layers and present the ASPP to robustly segment objects throughout many scales. DeepLab V3+ \cite{chen_deeplabv3plus} boosts the DeepLab family by adopting an encoder-decoder structure. More recent methods, such as OCR \cite{yuan_ocr} in conjunction with HRNetV2 \cite{wang_hrnetv2} or SegFormer \cite{xie_segformer}, have demonstrated the effectiveness of Transformers for Semantic Segmentation.
All of the above approaches seek to acquire semantics and global context at numerous scales, or in an encoder-decoder fashion, focusing mostly on autonomous driving scenarios. However, agricultural aerial imagery poses challenges that are not addressed by such architectures. With respect to these prior works, to better process the aerial data we guide the network to learn semantic representations that are invariant to the visual distortion and changes of orientation typical in the top-down perspective.

\subsection{Aerial Semantic Segmentation}
In aerial and remote sensing, the target environment of the semantic segmentation may vary greatly, from urban areas \cite{ref_beyond_rgb, ref_resunet_a,ref_seg_aerial_nogueira} to land cover \cite{ref_landcover_ai,ref_loveda,ref_deepglobe} and agricultural scenarios \cite{ref_agri_cnn, ref_agrivis_switchnorm, ref_agrivision}. 
These various target environments are generally linked to different applications, each with some specific challenges or requirements. 
For example, in urban monitoring semantic segmentation is mainly used to identify infrastructures, such as  roads \cite{ref_road_extr} and buildings \cite{ref_building}. This requires using high-resolution imagery as input and sometimes also to consider changes in time \cite{ref_change_detection}.
In land cover tasks, the main challenges are the extreme difference in the size of different semantic categories and the stark visual differences across different domains. Recent solutions address the first problem using multi-level or multi-scale feature aggregation \cite{ref_multiscale_fusion_aerial}, and the latter resorting to Domain Adaptation approaches \cite{ref_loveda,ref_landcover_da}.
When it comes to agricultural scenarios, classical segmentation solutions are mostly based on vegetation indices such as the NDVI \cite{weier2000_ndvi}, but the current trend is to move away from these handcrafted indices and towards more robust computer vision techniques, such as automated fusion of multi-spectral data \cite{ref_ndvi_fusion} and precise crop segmentation \cite{ref_agri_seg}.
Indeed, agricultural aerial images are rarely limited to the visible spectrum and frequently include other bands, such as Near-Infrared (NIR). In deep learning literature, the most common solutions to jointly exploit RGB and NIR images are the duplication of input weights \cite{ref_agrivision, ref_seg_coinnet} or multi-modal approaches based on late or early fusion \cite{ref_multiscale_fusion_aerial, ref_agrivis_switchnorm}. Another peculiarity of the aerial data is the fact that the orientation of the camera is arbitrary and uncertain. Although this problem has been addressed in incremental learning \cite{ref_rotinv_kd} and in classification tasks \cite{ref_rotinv_clf}, none of the current solutions in Semantic Segmentation consider this issue.

%% file: Sections/3-Method.tex
\begin{figure*}[!t]
\begin{center}
\includegraphics[width=1.0\textwidth]{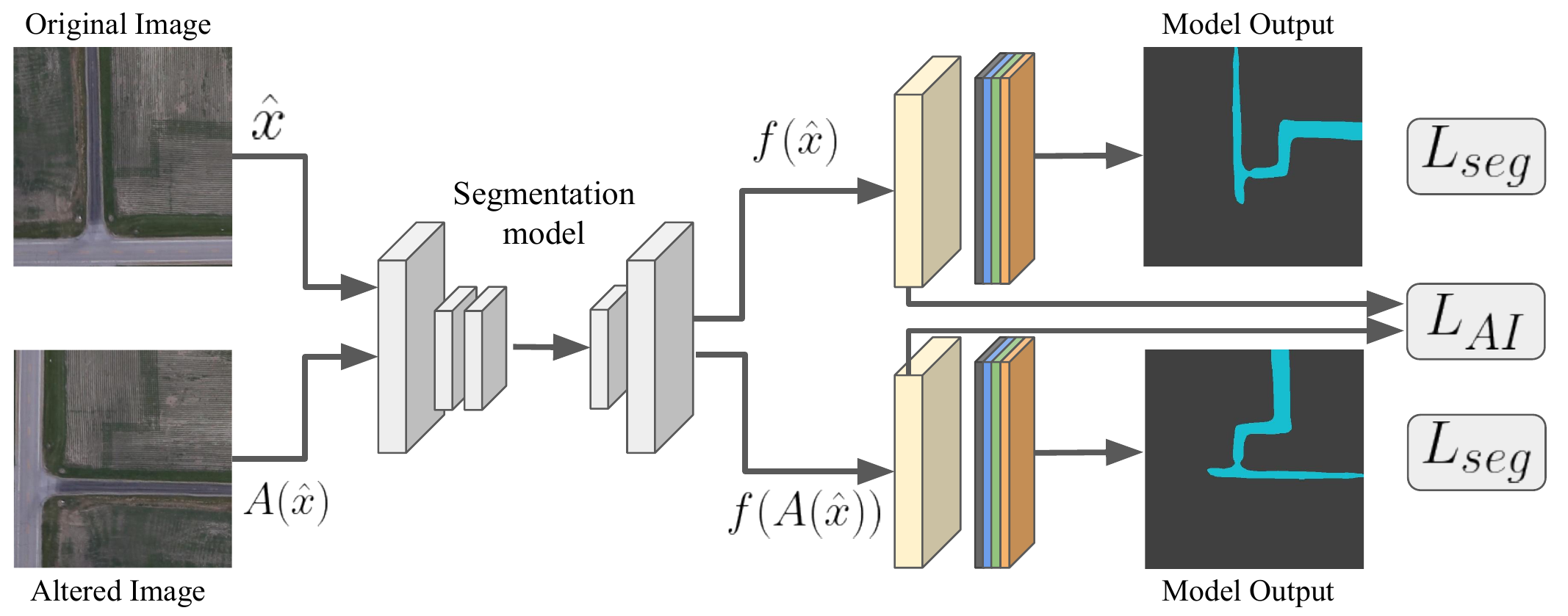}
\end{center}
\vspace{-15pt}
\caption{Illustration of the overall framework. The Adaptive Sampling picks a sample, and then an augmented version is generated. Both images are forwarded to the segmentation model, which computes the segmentation loss $L_{seg}$, whereas the $L_{AI}$ loss forces the model to extract the same features from both the original image and its transformed counterpart.}
\label{fig:framework}
\end{figure*}

\section{Method}
Our approach, depicted in \cref{fig:framework}, expands the SegFormer architecture \cite{xie_segformer} with two ideas. 
Firstly, by minimizing a loss that aligns the pixel embeddings generated by the Transformer network for the original image and for its augmented version, we promote the invariance of the learned semantic representations to photometric distortions and perspective changes that are typical in the aerial setting.
Secondly, we introduce an adaptive sampling mechanism to actively select the training samples using prior knowledge on class distribution and actual network confidence. In the rest of this section we first introduce the problem setting and then we detail these two mechanisms.

\subsection{Problem Setting}
We consider the problem of supervised semantic segmentation of agricultural aerial imagery, where during training we are provided with a collection of tuple $X = \{(x\in \mathcal{X}, y\in \mathcal{Y}, z\in \mathcal{Z})\}$,
with $\mathcal{X}$ being the set of RGB images, $\mathcal{Z}$ the set of Near-Infrared (NIR) images
and $\mathcal{Y}$ the set of semantic masks that associate to each pixel a class $c$ from a predefined set of semantic classes $\mathcal{C}$.
Additionally, we denote as $\mathcal{I}$ the set of pixels in each image and mask, and we define as $\hat{x} \in \mathcal{\hat{X}}$ the four-channel RGB-NIR image obtained by concatenating channel-wise $x$ and $z$. 

Our goal is to find a map $f_\theta: \mathcal{\hat{X}} \rightarrow \mathbb{R}^{|\mathcal{I}|\times|\mathcal{C}|}$, depending on a set of learnable parameters $\theta$, that assigns to each pixel of the RGB-NIR images an individual probability to belong to each semantic category in $\mathcal{C}$.
As a baseline, the parameters $\theta$ are optimized to minimize the standard cross-entropy loss $L_{seg}$:
\begin{equation}
    L_{\text{seg}}(\hat{x}, y) = - \frac{1}{|\mathcal{I}|} \sum_{i \in \mathcal{I}} \sum_{c \in \mathcal{C}}  y_i^c \log(p_i^c(\hat{x})),
    \label{eq:xe}
\end{equation}
where $p_i^c(\hat{x}) = f_\theta(\hat{x})[i,c]$ is the output of the model at a pixel $i$ for the class $c$, and $y_i^c$ represents the expected result for the same pixel and class.

\subsection{Augmentation Invariance}\label{sec:ai}
Current state-of-the-art frameworks are developed for the autonomous driving task and they suffer from performance degradation when applied to aerial data.
We identify a few factors that have a significant contribution to this degradation:
\begin{itemize}
    \item aerial images are not constrained to view the environment from a fixed perspective and, in particular, the camera orientation around the vertical axis is free;
    \item aerial images can display severe distortions due to the angle of the camera;
    \item there can be significant photometric shifts across different fields.
\end{itemize}

We propose a mechanism, called Augmentation Invariance (AI), which uses augmentations to guide the model to learn a mapping that is invariant to these shifts in perspective and appearance.
This mechanism works as follows. Given an input image $\hat{x}$, we extract the pixel-wise features $f_i(\hat{x})$ originating from the second-to-last layer of the SegFormer architecture at each iteration, explicitly skipping the last layer used for pixel-wise segmentation. 
Simultaneously, we transform a copy of $\hat{x}$ using both a random selection of geometric augmentations $A_g$ (\textit{horizontal flipping}, \textit{vertical flipping}, \textit{random rotation}), and a random photometric augmentation $A_p$ (\textit{color jitter}). Hereinafter, to simplify the notation, we denote the combination of both augmentations as $A_p \circ A_g = A$. The transformed image $A(\hat{x})$ is also passed through the model to extract the features $f_i(A(\hat{x}))$.
Finally, to make the model invariant to shifts in perspective and appearance we impose that the features extracted from the original image $\hat{x}$ are coincident with the features extracted from the transformed image $A(x)$, after reversing the geometric augmentation.
We achieve this with a pixel-wise mean squared error loss $L_{\text{AI}}$, which is defined as
\begin{equation}
    L_{\text{AI}}(\hat{x}, A(\hat{x})) = \frac{1}{\mathcal{I}} \sum_{i\in\mathcal{I}}(f_i(\hat{x}) - A_g^{-1}\left( f_i(A(\hat{x})) \right) )^2
    \label{eq:ai}
\end{equation}
In \cref{eq:ai}, $A_g^{-1}$ indicates the inversion of the geometric augmentations performed on $\hat{x}$, which is critical to ensure that we compare the original and augmented features corresponding to the same pixel.

We also maintain the ground truth annotations of the augmented images, so that the same segmentation loss can also be applied to them. The total training loss, denoted by $L_{tot}$, can be summarised as follows:

\begin{equation}
    L_{\text{tot}} = L_{\text{seg}}(\hat{x}, y) + L_{\text{seg}}(A(\hat{x}),A_g(y)) + \lambda  L_{\text{AI}}(\hat{x}, A(\hat{x})),
    \label{eq:tot}
\end{equation}
where $A_g(y)$ denotes the same geometric transformation applied to the ground truth annotation $y$, and $\lambda$ is a modulating factor. 

We remark that the mechanism of augmentation invariance used here is different from a classical data augmentation, because we do not use photometric and geometric transformations just to extend the training dataset with examples not in the original data distribution. Rather, through the loss $L_{AI}$ we also use the original and the transformed images paired together, to give a stronger guidance to the training process.

\subsection{Adaptive Sampling}\label{sec:as}
Despite the difficulty of avoiding learning bias from data, one of the primary issues in semantic aerial data is the strong distribution imbalance of semantic classes, with some that appear rarely and others that are extremely frequent. To address this issue, we use an Adaptive Sampling (AS) approach that works together with the Augmentation Invariance. At each iteration, the current sample of images required to train the network is selected according to two pieces of information: the global, pixel-wise class distribution, and the class-wise network confidence. In doing so, the data sampler will dynamically choose images giving priority to those whose categories appear with low frequency and for which the network has the least confidence.
Formally, the AS samples a class $c$ with an adaptive probability $AS_c$ defined as: 
\begin{equation}
    AS_c = \sigma((1 - dist * conf)^\gamma),
    \label{equation:as_1}
\end{equation}
where $dist$ is an array that represent the classes distribution, $conf$ represents the actual class-wise network confidence, $\sigma$ is a min-max normalization function and $\gamma$ is a relaxation parameter.
Once a semantic category $c$ has been chosen given this dynamically updated probability, an image is picked randomly from a subset of data $X_c$ that contains this class $c$. 
To compute \cref{equation:as_1} we use the following definitions of \textit{dist} and \textit{conf}:

\myparagraph{Class Distribution \textit{dist}.} Given that we are working in a supervised environment, we can compute a fixed, static distribution estimate as the amount of pixels for each semantic class $c\in \mathcal{C}$ only once, as a preprocessing step.
This array, which reflects the distribution of the classes, is normalized in the range $[0,1]$ and termed \textit{dist}. As a normalization step, we maintain the min-max normalization function.
In addition, for each class $c$, we keep track of the subset of images $X_c$ in which that category is represented.

\myparagraph{Network Confidence \textit{conf}.} We compute the network confidence for each class during training and store the result in an array with size $|C|$, named $conf$. At each iteration step $t$, the pixel-wise Softmax probabilities are computed on the current batch of prediction logits. The mean confidence value for each class $c$ is then derived from the available ground truth labels, by averaging pixels belonging to such category. Lastly, the actual network confidence is computed as the exponential moving average of the prior confidence at step $t-1$:
\begin{equation}
    conf_t = \alpha conf_{t-1} + (1-\alpha)conf_t,
    \label{equation:as_2}
\end{equation}
where $\alpha$ represents a smoothing factor.

%% file: Sections/4-Experiments.tex
\section{Experiments}

\subsection{Dataset and Metric}
We assess the performance of our approach considering the evaluation protocol described in Agriculture-Vision  \cite{ref_agrivision}. Due to the unavailability of the test set, we measure performances on the provided validation set. We conduct two sets of experiments: the first uses only RGB images for training and testing, while the second exploits NIR data in conjunction with the RGB images.

\input{Tables/rgb_exps}
\input{Tables/nirrgb_exps}

\myparagraph{Dataset.} Agriculture-Vision is a large-scale aerial farmland imagery collection for semantic segmentation of agricultural patterns. The dataset is made up of images collected from 3432 farmlands around the United States. It consists of 56944 RGB training images, which have been semantically labeled with 9 different categories. The validation set comprises instead 18334 images. In addition, for each training and validation sample, the NIR channel is also provided. Images are already provided in tiled format, 512 × 512 pixels each.

\myparagraph{Metric.} Following previous works, we adopt the standard average Intersection over Union metric (mIoU) \cite{pascal_iou} to measure the performance in all the experiments presented in the following sections.

\subsection{Implementation Details}

\myparagraph{Architecture.} The segmentation module at the base of our method is the SegFormer architecture \cite{xie_segformer}, using a MiT-B5 encoder pretrained on ImageNet-1k as backbone.

\myparagraph{Baselines.}
We compare our method against multiple baselines, taking into account the majority of state-of-the-art semantic segmentation techniques reported in the literature. The first model we examine is the FCN \cite{long_fcn}. We conduct experiments using the DeepLab V3 \cite{chen_deeplabv3} and DeepLab V3+ \cite{chen_deeplabv3plus} models from the DeepLab family. To compare with multi-scale techniques, we consider the FPN \cite{lin_fpn}, UperNet \cite{xiao_upernet}, and PSPNet \cite{zhao_pspnet}. All of these models are trained using ResNet-50 pretrained on ImageNet as the backbone. We then report results for the HRNetV2 method \cite{wang_hrnetv2} and its extension with transformer HRNetV2+OCR \cite{yuan_ocr}. HRNetV2-W18 pretrained on ImageNet serves as the backbone for both of them. Lastly, we examine the SegFormer architecture \cite{xie_segformer}, exploiting the standard pretrained encoder MiT-B5 with ImageNet weights, as it represents the baseline for our approach.

\myparagraph{Training.}
To develop our framework and reproduce all the baselines, we leverage the \textit{mmsegmentation} \cite{mmseg2020} framework, which is based on PyTorch. We train every configuration on two NVIDIA Tesla v100 GPUs with 16GB of RAM each. In terms of dataset augmentation, we employ random resizing with ratio in range (1.0, 2.0), random horizontal and vertical flipping, and random crops resized to 512x512 during training. Considering the evaluation pipeline, we perform inferences on raw data with no further preprocessing. We train all of the baselines and our model for 80k iterations using the AdamW optimizer. The learning rate is set to $6 \times 10^{-5}$, the weight decay to 0.01 and the betas are set to (0.9, 0.999). We use a \textit{poly} learning rate decay with a factor of 1.0 and an initial linear warm-up for 1500 iterations. We do not use class-balanced loss or OHEM approaches as in SegFormer \cite{xie_segformer}. When training using NIR data, we expand the network input to four channels by doubling the input weights of the red channel.

For the Augmentation Invariance (AI) variants, we further alter the available images using horizontal and vertical flipping, random rotation from 0° to 360° with a step of 90°, photometric and perspective distortion with a strength of 0.1. The probability for each transform is set to 0.5. We set the value of $\lambda$ in Eq. \cref{eq:ai} to 0.75 (see \cref{sec:ablation_lambda} for additional details).

\input{Tables/ablation_comp} 
\input{Tables/ablation_lambda}

Considering a hyperparameters search that compared the following values for $\gamma=\{1,2,4,6\}$ and for $\alpha=\{0.75, 0.85, 0.90, 0.968, 0.99\}$ on both settings, we set $\gamma=4$ on Eq. \cref{equation:as_1}, and $\alpha=0.968$ on Eq. \cref{equation:as_2}.

\subsection{Results}
\myparagraph{RGB.}
The results for this set of experiments are reported in \cref{table:rgb_exps}. The results confirm the difficulty of the task, as the averaged mIoU reaches $23.03\%$ when all the baseline approaches minus Transformer-based architectures are considered, while it increases to an average value of $32.68\%$ when OCR and SegFormer are added. With a mIoU of $21.67\%$, UperNet is the least performing approach. Despite this, it is one of the best in segmenting underrepresented classes, such as \textit{double plant}, \textit{waterways} or \textit{weed cluster}, as it is meant to capture multi-scale information. When using Transformer architectures, far better results can be observed, with a $30.07\%$ mIoU using the HRNetV2+OCR technique and even $44.96\%$ when using SegFormer; the improvement over UperNet is $+8.4\% $ and $+23.29\%$, respectively.

The majority of these strategies is designed for the autonomous driving domain, without considering the specific challenges intrinsic in aerial data. The introduction of our Augmentation Invariance and Adaptive Sampling results in a substantial boost in performance among almost all the semantic classes, especially on the underrepresented ones like \textit{double plant} or \textit{endrow}, yielding a total mIoU of $46.41\%$, and an improvement of $+24.74\%$ over the least performing UperNet and of $+1.45\%$ over the SegFormer architecture.

\myparagraph{NIR-RGB.} As expected, when using the additional Near-Infrared data provided by the Agriculture-Vision dataset, we observe performance improvements on all the baselines. The results are summarized in \cref{table:nirrgb_exps}. The average performance obtained from all baselines without considering transformers is $26.76\%$ in terms of mIoU, while it reaches $35.85\%$ on average when the Transformer-based architectures are also considered. Compared to the setting with RGB images only, the measured improvement reaches $+3.73\%$ and $+3.17\%$, respectively. This demonstrates how NIR infrared data enhances the whole training method by adding value and knowledge, in agreement with the literature \cite{ref_agrivis_switchnorm}. With a mIoU of $23.93\%$, the FCN architecture achieves the lowest score, while the SegFormer architecture represents again the best performing approach among the baselines with a mIoU of $46.50\%$. The overall improvement in comparison to FCN is $+22.57\%$.

In this set of experiments, our solution appears to be successful, achieving the best performance among all considered approaches, with a mIoU of $ 49.04\%$. AI and AS improve the performance in all the semantic classes, with some outliers in underrepresented ones, such as \textit{double plant}, which gains a $+ 27.92\%$, \textit{endrow}, which gains a $+ 24.03\%$, \textit{planter skip}, which gains a $+ 32.04\%$ and \textit{waterways}, which gains a $+43.24$ w.r.t. the least performant FCN. The overall improvement in comparison to FCN that AI and AS allow to reach is of $+25.11\% $. These results and the qualitatives in \cref{fig:qualitatives} confirm the validity and effectiveness of our solution in dealing with the primary challenges raised by this task.

%% file: Tables/rgb_exps.tex
\begin{table*}[!ht]
\centering
\begin{adjustbox}{width=1.0\textwidth}
\begin{tabular}{l|ccccccccc|l}
\multirow{2}{*}{Method} & \multicolumn{9}{c|}{Semantic Classes IoU}                                                                                                                                                                                                                                                                 & \multirow{2}{*}{mIoU}  \\
                                          & \multicolumn{1}{l}{Background} & \multicolumn{1}{l}{Double Plant} & \multicolumn{1}{l}{Drydown} & \multicolumn{1}{l}{Endrow} & \multicolumn{1}{l}{Nutrient Deficiency} & \multicolumn{1}{l}{Planter Skip} & \multicolumn{1}{l}{Water} & \multicolumn{1}{l}{Waterways} & \multicolumn{1}{l|}{Weed Cluster} &                        \\ 
\hline
FCN                                       & 69.99                          & 16.91                            & 45.55                       & 0.18                       & 13.66                                   & 6.62                             & 42.27                     & 0.52                          & 8.50                              & 22.91                  \\
DeepLab V3                                & 66.27                          & 17.01                            & 40.64                       & 9.46                       & 16.40                                   & 10.04                            & 17.06                     & 12.29                         & 9.97                              & 22.13                  \\
DeepLab V3+                               & 68.55                          & 16.31                            & 46.36                       & 6.46                       & 16.05                                   & 4.56                             & 16.61                     & 19.10                         & 13.89                             & 23.10                  \\
UperNet                                   & 65.84                          & 15.79                            & 38.03                       & 10.12                      & 17.31                                   & 11.09                            & 4.47                      & 15.45                         & 16.94                             & 21.67                  \\
SFPN                                      & 69.65                          & 10.61                            & 49.49                       & 2.70                       & 11.46                                   & 4.80                             & 35.68                     & 9.89                          & 11.16                             & 22.83                  \\
PSPNet                                    & 68.11                          & 16.93                            & 45.77                       & 4.89                       & 18.99                                   & 8.54                             & 11.31                     & 17.64                         & 17.20                             & 23.26                  \\
HRNetV2                                  & 71.21                          & 16.81                            & 55.10                       & 5.22                       & 18.63                                   & 13.26                            & 13.03                     & 21.23                         & 14.07                             & 25.39                  \\
HRNetV2+OCR                           & 72.42                          & 19.46                            & 56.79                       & 12.31                      & 17.30                                   & 21.31                            & 28.36                     & 24.62                         & 18.05                             & 30.07                  \\
SegFormer                                 & 74.93                          & 33.19                            & \textbf{59.65}                       & 18.28                      & \textbf{31.64}                                   & 39.20                            & 77.97                     & \textbf{41.45}                         & 28.31                             & 44.96                  \\ 
\hline
\textbf{Ours}   & \textbf{75.47} & \textbf{36.97}  & 58.49  & \textbf{22.69}                          & 31.29  & \textbf{41.39}  & \textbf{80.23}                         & 40.07        & \textbf{30.42}    & \textbf{46.41}                     
\end{tabular}
\end{adjustbox}
\caption{Experiments using RGB images for training and testing on the Agriculture Vision dataset.}
\label{table:rgb_exps}
\end{table*}

%% file: Tables/nirrgb_exps.tex
\begin{table*}[!ht]
\centering
\begin{adjustbox}{width=1.0\textwidth}
\begin{tabular}{l|ccccccccc|l}
\multirow{2}{*}{Method} & \multicolumn{9}{c|}{Semantic Classes IoU}                                                                & \multirow{2}{*}{mIoU}  \\
                        & Background & Double Plant & Drydown & Endrow & Nutrient Deficiency & Planter Skip & Water & Waterways & Weed Cluster &                        \\ 
\hline
FCN                     & 68.35      & 9.40         & 47.57   & 0.54   & 15.16               & 9.97         & 53.74 & 0.47      & 10.17        & 23.93                  \\
DeepLab V3              & 69.03      & 19.97        & 43.94   & 5.85   & 23.98               & 17.86        & 46.74 & 29.03     & 11.36        & 29.75                  \\
DeepLab V3+             & 68.29      & 17.18        & 48.07   & 7.48   & 24.17               & 19.57        & 19.43 & 24.58     & 13.22        & 26.89                  \\
UperNet                 & 67.43      & 15.63        & 36.40   & 10.73  & 20.37               & 14.57        & 34.21 & 25.28     & 14.54        & 26.57                  \\
SFPN                    & 68.69      & 5.99         & 48.71   & 0.18   & 22.74               & 17.21        & 44.50 & 18.30     & 12.79        & 26.57                  \\
PSPNet                  & 66.92      & 17.73        & 29.87   & 10.24  & 28.01               & 18.66        & 13.90 & 29.83     & 11.99        & 25.24                  \\
HRNetV2                & 71.28      & 16.99        & 54.30   & 4.52   & 27.90               & 15.74        & 21.66 & 25.47     & 17.88        & 28.42                  \\
HRNetV2+OCR         & 72.60      & 17.98        & 56.69   & 11.97  & 27.91               & 23.79        & 48.99 & 27.73     & 22.06        & 34.42                  \\
SegFormer               & 76.17      & 33.63        & 58.96   & 18.92  & 40.57               & 38.93        & 80.56 & 42.85     & 27.88        & 46.50                  \\ 
\hline
\textbf{Ours} &  \textbf{76.19}   & \textbf{37.32} & \textbf{61.75} & \textbf{24.57}  & \textbf{42.75}  & \textbf{42.01}  & \textbf{81.32} & \textbf{43.71} & \textbf{31.75}      & \textbf{49.04}               
\end{tabular}
\end{adjustbox}
\caption{Experiments using the combination of NIR data and RGB images for training and testing on the Agriculture Vision dataset.}
\label{table:nirrgb_exps}
\vspace{-5pt}
\end{table*}

%% file: Tables/ablation_comp.tex
\begin{table*}[!ht]
\centering
\begin{adjustbox}{width=1.0\textwidth}
\begin{tabular}{l|ccccccccc|l}
\multirow{2}{*}{Components ~} & \multicolumn{9}{c|}{Semantic Classes IoU ~ ~ ~ ~ ~ ~}                                                                & \multirow{2}{*}{mIoU}  \\
                              & Background & Double Plant & Drydown & Endrow & Nutrient Deficiency & Planter Skip & Water & Waterways & Weed Cluster &                        \\ 
\hline
SegFormer                     & 76.17      & 33.63        & 58.96   & 18.92  & 40.57               & 38.93        & 80.56 & 42.85     & 27.88        & 46.50                  \\
SegFormer + AI                & \underline{76.62}      & 35.26        & 61.24   & 20.74  & \underline{43.45}               & \underline{43.49}        & 80.41 & \underline{45.10}     & \underline{33.12}        & 48.82                  \\
SegFormer + AS                & 75.89          & 35.86            & 59.23       & 22.5      & 41.25                   & 40.72            & 77.98     & 40.85         & 30.99            & 47.25                      \\
SegFormer + AI + AS     &    76.19   & \underline{37.32} & \underline{61.75}      & \underline{24.57}      &    42.75                & 42.01             &  \underline{81.32}  &    43.71      &    31.75         & \textbf{49.04}                      
\end{tabular}
\end{adjustbox}
\caption{Ablation study showing the effectiveness of the AI and AS components on the NIR-RGB setting.}
\label{table:ab_comp}
\end{table*}

%% file: Tables/ablation_lambda.tex
\begin{table*}[!ht]
\centering
\begin{adjustbox}{width=1.0\textwidth}
\begin{tabular}{l|ccccccccc|l}
\multirow{2}{*}{$\lambda$} & \multicolumn{9}{c|}{Semantic Classes IoU}                                                                & \multirow{2}{*}{mIoU}  \\
                        & Background & Double Plant & Drydown & Endrow & Nutrient Deficiency & Planter Skip & Water & Waterways & Weed Cluster &                        \\ 
\hline
0.1                     & 76.60      & 33.92        & 60.24   & 18.84  & 41.92               & 41.28        & \underline{82.23} & 42.45     & 31.70        & 47.69                  \\
0.25                    & 76.54      & 35.26        & 60.70   & 20.55  & 42.22               & \underline{43.84}        & 80.60 & 43.16     & \underline{33.25}        & 48.46                  \\
0.5                     & 76.48      & \underline{35.79}        & 59.71   & 20.34  & 42.65               & 40.03        & 81.12 & 44.52     & 32.00        & 48.07                  \\
0.75                    & \underline{76.62}      & 35.26        & \underline{61.24}   & \underline{20.74}  & \underline{43.45}               & 43.49        & 80.41 & \underline{45.10}     & 33.12        & \textbf{48.82}                  \\
1                       & 76.57      & 34.42        & 60.25   & 20.32  & 41.95               & 40.03        & 82.14 & 43.51     & 32.22        & 47.93                 
\end{tabular}
\end{adjustbox}
\caption{Ablation study on the influence of $\lambda$ on the NIR-RGB setting.}
\label{table:ab_lambda}
\end{table*}

%% file: Sections/5-Ablation.tex
\begin{figure*}[!ht]
\begin{center}
\includegraphics[width=0.96\textwidth]{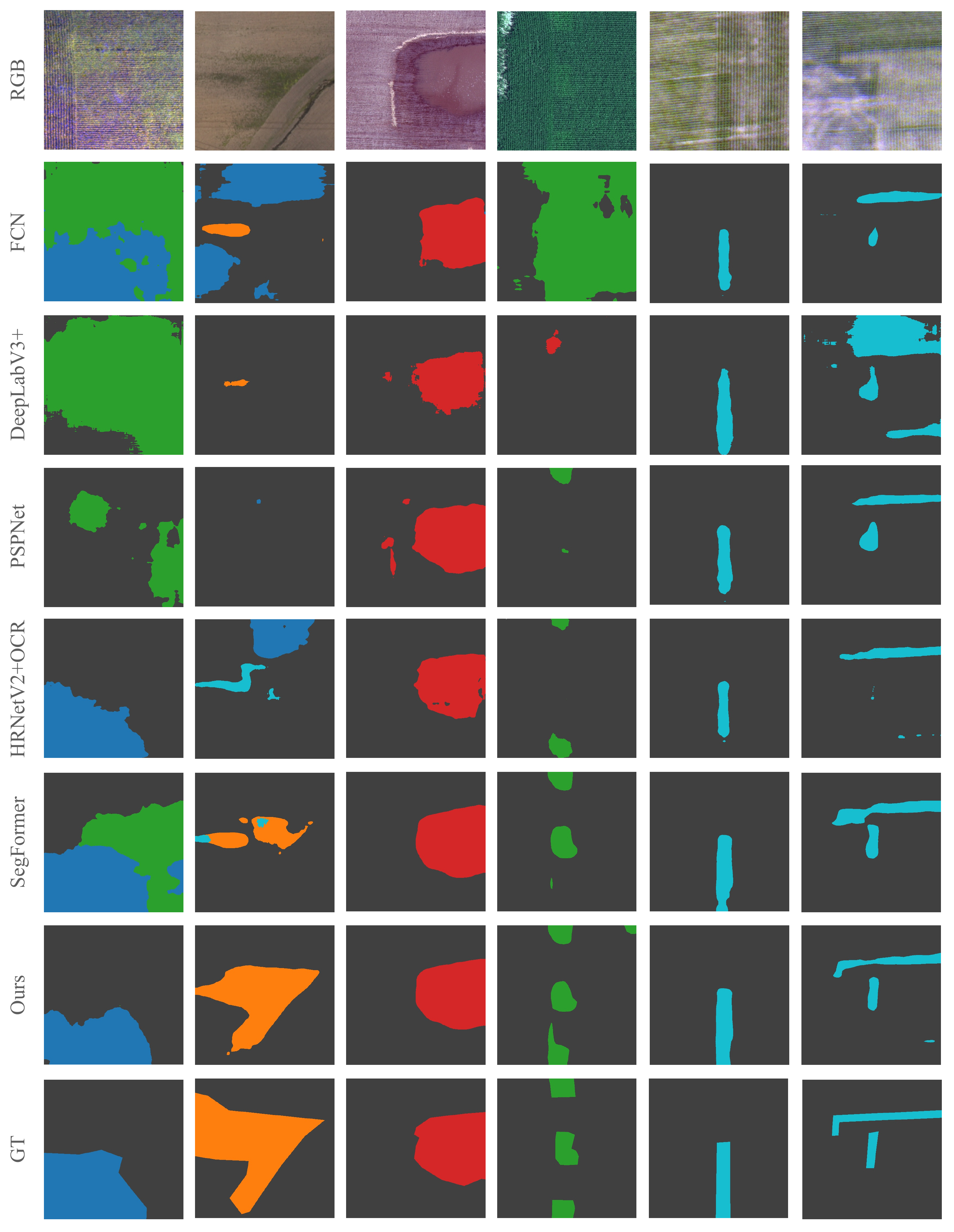}
\end{center}
\vspace{-15pt}
\caption{Qualitative results on the validation set of the Agriculture-Vision dataset.}
\label{fig:qualitatives}
\end{figure*}

\section{Ablation study}

\subsection{Contribution of each component} \label{sec:ablation_comp}
In this section, we assess how each proposed component contributes to the overall performance of our method. We investigate four distinct cases: (a) the SegFormer framework, (b) the introduction of our Augmentation Invariance (AI), (c) the introduction of our Adaptive Sampling (AS) technique, and lastly (d) the entire framework, which includes both AI and AS.
We report the results in \cref{table:ab_comp}. This table shows how the Augmentation Invariance is critical for providing a boost to the overall framework, thus confirming our conjecture about the specific challenges in agricultural aerial imagery. The achieved boost is $+2.32 \% $ in comparison to the baseline architecture. The addition of AS boosts the simple SegFormer design, delivering state-of-the-art results and highlighting the need to address the semantic class imbalance.
The combination of AI and AS provides a further improvement, particularly on underrepresented classes, e.g. \textit{double plant} rise of $+3.69\%$ and \textit{endrow} rise of $+5.65\%$ w.r.t. SegFormer, and of $+2.06 \% $ and $+3.83 \% $ w.r.t. AI, respectively.

\subsection{Ablation on $\lambda$} \label{sec:ablation_lambda}
The $\lambda$ hyperparameter is required to determine the intensity of the Augmentation Invariance (AI) loss. The following values of $\lambda$ are being compared: 0.1, 0.25, 0.5, 0.75, and 1.0. We run the experiments using the NIR-RGB protocol, without applying the Adaptive Sampling, and we report the outcomes in \cref{table:ab_lambda}. The best results are obtained when $\lambda=0.75$. 
Even though $\lambda=0.1$ yields the lowest performance, with a difference of $1.13\%$ when compared to $\lambda=0.75$, the achieved score can still be considered state of the art on its own. In conclusion, even when using sub-optimal hyperparameters, our AI outperforms all the baselines, highlighting the effectiveness of the approach.

%% file: Sections/6-Conclusion.tex
\section{Conclusions}
\myparagraph{Limitations.} 
When we apply Adaptive Sampling to the Agriculture-Vision dataset, we see a modest drop in performance on some categories, such as \textit{planter skip} or \textit{waterways}. This is because the difference in absolute pixel counts between these categories (excluding the background class) does not appear to be extremely significant, limiting the influence of the AS technique on the final outcome.
Moreover, we note that the adopted training configuration might not be optimal for this setting, therefore hyperparameter changes, such as a higher iteration count, may limit or completely solve these issues.

\myparagraph{Conclusion.}
In this paper we address the problem of Semantic Segmentation for agricultural aerial images. Aside from the standard issues in semantic segmentation, delineating patterns in aerial imagery poses additional challenges such as how to leverage the additional multi-modal data that comes with the visible spectrum, the imbalance in class-wise pixel distribution, and the changes in point of view. 
We offer two approaches to address these challenges in an end-to-end trainable framework: an Augmentation Invariance solution that forces the model to learn semantic representations that are invariant to the point-of-view shifts typical in aerial imagery, and an Adaptive Sampling solution that addresses the problem of class imbalance by actively sampling  the training images based on their class-wise pixel distribution and the current network confidence.

We propose a comprehensive series of experiments and ablation studies on the Agriculture-Vision dataset and we prove how our methods considerably increase the performance of the actual state-of-the-art models, especially on underrepresented classes.

Future research will look into the feasibility of developing a plug-and-play technique for bringing Augmentation Invariance and Adaptive Sampling to any segmentation backbone. Furthermore, we will examine the influence of various types of augmentations employed to force the model to be agnostic. We also intend to evaluate our method in a variety of contexts, including Domain Adaptation and Domain Generalization.

%% file: Main.bbl
\begin{thebibliography}{10}\itemsep=-1pt

\bibitem{idda}
E. {Alberti}, A. {Tavera}, C. {Masone}, and B. {Caputo}.
\newblock Idda: A large-scale multi-domain dataset for autonomous driving.
\newblock {\em IEEE Robotics and Automation Letters}, 5(4):5526--5533, 2020.

\bibitem{andrade_deforestation}
RB Andrade, GAOP Costa, GLA Mota, MX Ortega, RQ Feitosa, PJ Soto, and Christian
  Heipke.
\newblock Evaluation of semantic segmentation methods for deforestation
  detection in the amazon.
\newblock {\em ISPRS Archives; 43, B3}, 43(B3):1497--1505, 2020.

\bibitem{ref_rotinv_kd}
Edoardo Arnaudo, Fabio Cermelli, Antonio Tavera, Claudio Rossi, and Barbara
  Caputo.
\newblock A contrastive distillation approach for incremental semantic
  segmentation in aerial images.
\newblock {\em arXiv preprint arXiv:2112.03814}, 2021.

\bibitem{ref_beyond_rgb}
Nicolas Audebert, Bertrand {Le Saux}, and Sébastien Lefèvre.
\newblock Beyond rgb: Very high resolution urban remote sensing with multimodal
  deep networks.
\newblock {\em ISPRS Journ. Phot. Rem. Sens.}, 140:20--32, 2018.

\bibitem{ref_landcover_da}
Nadir Bengana and Janne Heikkil{\"a}.
\newblock Improving land cover segmentation across satellites using domain
  adaptation.
\newblock {\em IEEE Journal of Selected Topics in Applied Earth Observations
  and Remote Sensing}, 14:1399--1410, 2020.

\bibitem{ref_landcover_ai}
Adrian Boguszewski, Dominik Batorski, Natalia Ziemba-Jankowska, Tomasz
  Dziedzic, and Anna Zambrzycka.
\newblock Landcover. ai: Dataset for automatic mapping of buildings, woodlands,
  water and roads from aerial imagery.
\newblock In {\em Proceedings of the IEEE/CVF Conference on Computer Vision and
  Pattern Recognition}, pages 1102--1110, 2021.

\bibitem{chen_deeplabv2}
Liang-Chieh Chen, George Papandreou, Iasonas Kokkinos, Kevin~P. Murphy, and
  Alan~Loddon Yuille.
\newblock Deeplab: Semantic image segmentation with deep convolutional nets,
  atrous convolution, and fully connected crfs.
\newblock {\em IEEE Transactions on Pattern Analysis and Machine Intelligence},
  40:834--848, 2018.

\bibitem{chen_deeplabv3}
Liang-Chieh Chen, George Papandreou, Florian Schroff, and Hartwig Adam.
\newblock Rethinking atrous convolution for semantic image segmentation.
\newblock {\em ArXiv}, abs/1706.05587, 2017.

\bibitem{chen_deeplabv3plus}
Liang-Chieh Chen, Yukun Zhu, George Papandreou, Florian Schroff, and Hartwig
  Adam.
\newblock Encoder-decoder with atrous separable convolution for semantic image
  segmentation.
\newblock In {\em Proceedings of the European conference on computer vision
  (ECCV)}, pages 801--818, 2018.

\bibitem{ref_building}
Dominic Cheng, Renjie Liao, Sanja Fidler, and Raquel Urtasun.
\newblock Darnet: Deep active ray network for building segmentation.
\newblock In {\em Proceedings of the IEEE/CVF Conference on Computer Vision and
  Pattern Recognition}, pages 7431--7439, 2019.

\bibitem{ref_agrivision}
Mang~Tik Chiu, Xingqian Xu, Yunchao Wei, Zilong Huang, Alexander~G Schwing,
  Robert Brunner, Hrant Khachatrian, Hovnatan Karapetyan, Ivan Dozier, Greg
  Rose, et~al.
\newblock Agriculture-vision: A large aerial image database for agricultural
  pattern analysis.
\newblock In {\em Proceedings of the IEEE/CVF Conference on Computer Vision and
  Pattern Recognition}, pages 2828--2838, 2020.

\bibitem{mmseg2020}
MMSegmentation Contributors.
\newblock {MMSegmentation}: Openmmlab semantic segmentation toolbox and
  benchmark.
\newblock \url{https://github.com/open-mmlab/mmsegmentation}, 2020.

\bibitem{cityscapes}
Marius Cordts, Mohamed Omran, Sebastian Ramos, Timo Rehfeld, Markus Enzweiler,
  Rodrigo Benenson, Uwe Franke, Stefan Roth, and Bernt Schiele.
\newblock The cityscapes dataset for semantic urban scene understanding.
\newblock In {\em CVPR}, 2016.

\bibitem{ref_deepglobe}
Ilke Demir, Krzysztof Koperski, David Lindenbaum, Guan Pang, Jing Huang, Saikat
  Basu, Forest Hughes, Devis Tuia, and Ramesh Raskar.
\newblock Deepglobe 2018: A challenge to parse the earth through satellite
  images.
\newblock In {\em Proceedings of the IEEE Conference on Computer Vision and
  Pattern Recognition Workshops}, pages 172--181, 2018.

\bibitem{ref_resunet_a}
Foivos~I. Diakogiannis, François Waldner, Peter Caccetta, and Chen Wu.
\newblock Resunet-a: A deep learning framework for semantic segmentation of
  remotely sensed data.
\newblock {\em ISPRS Journ. Phot. Rem. Sens.}, 162:94--114, 2020.

\bibitem{pascal_iou}
Mark Everingham, S.~M.~Ali Eslami, Luc~Van Gool, Christopher K.~I. Williams,
  John~M. Winn, and Andrew Zisserman.
\newblock The pascal visual object classes challenge: A retrospective.
\newblock {\em International Journal of Computer Vision}, 111:98--136, 2014.

\bibitem{farasin_doublestep}
Alessandro Farasin, Luca Colomba, and Paolo Garza.
\newblock Double-step u-net: A deep learning-based approach for the estimation
  of wildfire damage severity through sentinel-2 satellite data.
\newblock {\em Applied Sciences}, 10(12):4332, 2020.

\bibitem{ref_agri_seg}
Mulham Fawakherji, Ali Youssef, Domenico Bloisi, Alberto Pretto, and Daniele
  Nardi.
\newblock Crop and weeds classification for precision agriculture using
  context-independent pixel-wise segmentation.
\newblock In {\em 2019 Third IEEE International Conference on Robotic Computing
  (IRC)}, pages 146--152. IEEE, 2019.

\bibitem{lin_fpn}
Tsung-Yi Lin, Piotr Doll{\'a}r, Ross~B. Girshick, Kaiming He, Bharath
  Hariharan, and Serge~J. Belongie.
\newblock Feature pyramid networks for object detection.
\newblock {\em 2017 IEEE Conference on Computer Vision and Pattern Recognition
  (CVPR)}, pages 936--944, 2017.

\bibitem{long_fcn}
J. Long, E. Shelhamer, and T. Darrell.
\newblock Fully convolutional networks for semantic segmentation.
\newblock In {\em IEEE Conf. Comput. Vis. Pattern Recog. (CVPR)}, 2015.

\bibitem{ref_change_detection}
Ning Lv, Chen Chen, Tie Qiu, and Arun~Kumar Sangaiah.
\newblock Deep learning and superpixel feature extraction based on contractive
  autoencoder for change detection in sar images.
\newblock {\em IEEE transactions on industrial informatics}, 14(12):5530--5538,
  2018.

\bibitem{ref_agri_cnn}
Andres Milioto, Philipp Lottes, and Cyrill Stachniss.
\newblock Real-time semantic segmentation of crop and weed for precision
  agriculture robots leveraging background knowledge in cnns.
\newblock In {\em 2018 IEEE international conference on robotics and automation
  (ICRA)}, pages 2229--2235. IEEE, 2018.

\bibitem{semseg_survey}
Shervin Minaee, Yuri~Y. Boykov, Fatih Porikli, Antonio~J Plaza, Nasser
  Kehtarnavaz, and Demetri Terzopoulos.
\newblock Image segmentation using deep learning: A survey.
\newblock {\em IEEE Transactions on Pattern Analysis and Machine Intelligence},
  pages 1--1, 2021.

\bibitem{ref_seg_aerial_nogueira}
Keiller Nogueira, Mauro Dalla~Mura, Jocelyn Chanussot, William~Robson Schwartz,
  and Jefersson~A. dos Santos.
\newblock Learning to semantically segment high-resolution remote sensing
  images.
\newblock In {\em Int. Conf. Pattern Recog.}, pages 3566--3571, 2016.

\bibitem{ref_seg_coinnet}
Bin Pan, Zhenwei Shi, Xia Xu, Tianyang Shi, Ning Zhang, and Xinzhong Zhu.
\newblock Coinnet: Copy initialization network for multispectral imagery
  semantic segmentation.
\newblock {\em IEEE Geos. Rem. Sens. Lett.}, 16(5):816--820, 2019.

\bibitem{ref_rotinv_clf}
Kunlun Qi, Chao Yang, Chuli Hu, Yonglin Shen, Shengyu Shen, and Huayi Wu.
\newblock Rotation invariance regularization for remote sensing image scene
  classification with convolutional neural networks.
\newblock {\em Rem. Sens.}, 13(4), 2021.

\bibitem{ronneberger_unet}
Olaf Ronneberger, Philipp Fischer, and Thomas Brox.
\newblock U-net: Convolutional networks for biomedical image segmentation.
\newblock In Nassir Navab, Joachim Hornegger, William~M. Wells, and
  Alejandro~F. Frangi, editors, {\em Medical Image Computing and
  Computer-Assisted Intervention (MICCAI)}, pages 234--241, 2015.

\bibitem{ref_ndvi_fusion}
Hao Sheng, Xiao Chen, Jingyi Su, Ram Rajagopal, and Andrew Ng.
\newblock Effective data fusion with generalized vegetation index: Evidence
  from land cover segmentation in agriculture.
\newblock In {\em Proceedings of the IEEE/CVF Conference on Computer Vision and
  Pattern Recognition Workshops}, pages 60--61, 2020.

\bibitem{sun_hrnetv1}
Ke Sun, Yang Zhao, Borui Jiang, Tianheng Cheng, Bin Xiao, Dong Liu, Yadong Mu,
  Xinggang Wang, Wenyu Liu, and Jingdong Wang.
\newblock High-resolution representations for labeling pixels and regions.
\newblock {\em CoRR}, abs/1904.04514, 2019.

\bibitem{Tavera_2022_WACV}
Antonio Tavera, Fabio Cermelli, Carlo Masone, and Barbara Caputo.
\newblock Pixel-by-pixel cross-domain alignment for few-shot semantic
  segmentation.
\newblock In {\em Proceedings of the IEEE/CVF Winter Conference on Applications
  of Computer Vision (WACV)}, pages 1626--1635, January 2022.

\bibitem{wang_hrnetv2}
Jingdong Wang, Ke Sun, Tianheng Cheng, Borui Jiang, Chaorui Deng, Yang Zhao,
  Dong Liu, Yadong Mu, Mingkui Tan, Xinggang Wang, Wenyu Liu, and Bin Xiao.
\newblock Deep high-resolution representation learning for visual recognition.
\newblock {\em TPAMI}, 2019.

\bibitem{ref_loveda}
Junjue Wang, Zhuo Zheng, Ailong Ma, Xiaoyan Lu, and Yanfei Zhong.
\newblock Loveda: A remote sensing land-cover dataset for domain adaptive
  semantic segmentation.
\newblock {\em arXiv preprint arXiv:2110.08733}, 2021.

\bibitem{weier2000_ndvi}
John Weier and David Herring.
\newblock Measuring vegetation (ndvi \& evi).
\newblock {\em NASA Earth Observatory}, 20, 2000.

\bibitem{xiao_upernet}
Tete Xiao, Yingcheng Liu, Bolei Zhou, Yuning Jiang, and Jian Sun.
\newblock Unified perceptual parsing for scene understanding.
\newblock In {\em Proceedings of the European Conference on Computer Vision
  (ECCV)}, pages 418--434, 2018.

\bibitem{xie_segformer}
Enze Xie, Wenhai Wang, Zhiding Yu, Anima Anandkumar, Jose~M Alvarez, and Ping
  Luo.
\newblock Segformer: Simple and efficient design for semantic segmentation with
  transformers.
\newblock {\em Advances in Neural Information Processing Systems}, 34, 2021.

\bibitem{ref_agrivis_switchnorm}
S. Yang, S. Yu, B. Zhao, and Y. Wang.
\newblock Reducing the feature divergence of rgb and near-infrared images using
  switchable normalization.
\newblock In {\em IEEE Conf. Comput. Vis. Pattern Recog. Work.}, pages
  206--211, jun 2020.

\bibitem{ref_multiscale_fusion_aerial}
Qinglie Yuan, Helmi Zulhaidi~Mohd Shafri, Aidi~Hizami Alias, and Shaiful
  Jahari~bin Hashim.
\newblock Multiscale semantic feature optimization and fusion network for
  building extraction using high-resolution aerial images and lidar data.
\newblock {\em Rem. Sens.}, 13(13), 2021.

\bibitem{yuan_ocr}
Yuhui Yuan, Xiaokang Chen, Xilin Chen, and Jingdong Wang.
\newblock Segmentation transformer: Object-contextual representations for
  semantic segmentation.
\newblock {\em arXiv preprint arXiv:1909.11065}, 2019.

\bibitem{ref_road_extr}
Zhengxin Zhang, Qingjie Liu, and Yunhong Wang.
\newblock Road extraction by deep residual u-net.
\newblock {\em IEEE Geoscience and Remote Sensing Letters}, 15(5):749--753,
  2018.

\bibitem{zhao_pspnet}
Hengshuang Zhao, Jianping Shi, Xiaojuan Qi, Xiaogang Wang, and Jiaya Jia.
\newblock Pyramid scene parsing network.
\newblock {\em 2017 IEEE Conference on Computer Vision and Pattern Recognition
  (CVPR)}, pages 6230--6239, 2017.

\end{thebibliography}
